\pdfoutput=1

\documentclass[11pt]{article}

\usepackage{acl}
\usepackage{subfigure}
\usepackage{times}
\usepackage{latexsym}
\usepackage{amsmath}
\usepackage{amssymb}
\usepackage{colortbl}
\usepackage[T1]{fontenc}

\usepackage[utf8]{inputenc}

\usepackage{microtype}

\usepackage{inconsolata}
\usepackage{array}
\usepackage{graphicx}
\usepackage{booktabs}
\usepackage{multirow}
\usepackage{multicol}
\usepackage{algorithm}
\usepackage{algorithmic}

\usepackage{bbding}
\usepackage{pifont}
\usepackage{wasysym}
\usepackage{utfsym}
\usepackage{fontawesome}
\usepackage{mathtools}

\title{Towards Stepwise Domain Knowledge-Driven Reasoning Optimization\\ and Reflection Improvement}

\author{\textbf{Chengyuan Liu\textsuperscript{1,2}\footnotemark[2],\quad Shihang Wang\textsuperscript{2},\quad Lizhi Qing\textsuperscript{2},\quad Kaisong Song\textsuperscript{2},}\\ \textbf{Junjie Cao\textsuperscript{2},\quad Jun Lin\textsuperscript{2},\quad Ji Zhang\textsuperscript{2},\quad Ang Li\textsuperscript{1},\quad Kun Kuang\textsuperscript{1,3}\footnotemark[1],\quad Fei Wu\textsuperscript{1}}\\
\textit{\small \{liucy1,wufei,kunkuang\}@zju.edu.cn}, \\ \textit{\small \{wangshihang.wsh,yekai.qlz, kaisong.sks,junjie.junjiecao,linjun.lj,zj122146\}@alibaba-inc.com}
\\
\textit{\small \textsuperscript{1}College of Computer Science and Technology, Zhejiang University,}\\
\textit{\small \textsuperscript{2}Tongyi Lab, Alibaba Group,}
\textit{\small \textsuperscript{3}Law\&AI Lab, Zhejiang University}\\
}

\makeatletter
\renewcommand{\maketag@@@}[1]{\hbox{\m@th\normalsize\normalfont#1}}%
\makeatother

\begin{document}
\maketitle
\renewcommand{\thefootnote}{\fnsymbol{footnote}}
\footnotetext[1]{Corresponding author.}
\footnotetext[2]{This work was done when Chengyuan Liu interned at Alibaba.}
\renewcommand*{\thefootnote}{\arabic{footnote}}
\begin{abstract}
Recently, stepwise supervision on Chain of Thoughts (CoTs) presents an enhancement on the logical reasoning tasks such as coding and math, with the help of Monte Carlo Tree Search (MCTS). However, its contribution to tasks requiring domain-specific expertise and knowledge remains unexplored. 
Motivated by the interest, we identify several potential challenges of vanilla MCTS within this context, and propose the framework of Stepwise Domain Knowledge-Driven Reasoning Optimization, employing the MCTS algorithm to develop step-level supervision for problems that require essential comprehension, reasoning, and specialized knowledge.  Additionally, we also introduce the Preference Optimization towards Reflection Paths, which iteratively learns self-reflection on the reasoning thoughts from better perspectives. We have conducted extensive experiments to evaluate the advantage of the methodologies. Empirical results demonstrate the effectiveness on various legal-domain problems. We also report a diverse set of valuable findings, hoping to encourage the enthusiasm to the research of domain-specific LLMs and MCTS.

\end{abstract}

\section{Introduction}

Chains of Thought (CoT) facilitates logical reasoning ability by explicitly detailing the thought process step-by-step, thereby improving accuracy on tasks such as coding and math \cite{wei2023chainofthoughtpromptingelicitsreasoning,narangself,zhang2023automatic}. Recently, Monte Carlo Tree Search (MCTS) algorithm brings further enhancement to the logic reasoning of Large Language Models (LLMs) by providing fine-grained supervision on each step of the solution \cite{chen2024alphamathzeroprocesssupervision,svpo,tian2024selfimprovementllmsimaginationsearching,zhang2024rest}.

\begin{figure}
    \centering
    \includegraphics[width=\linewidth]{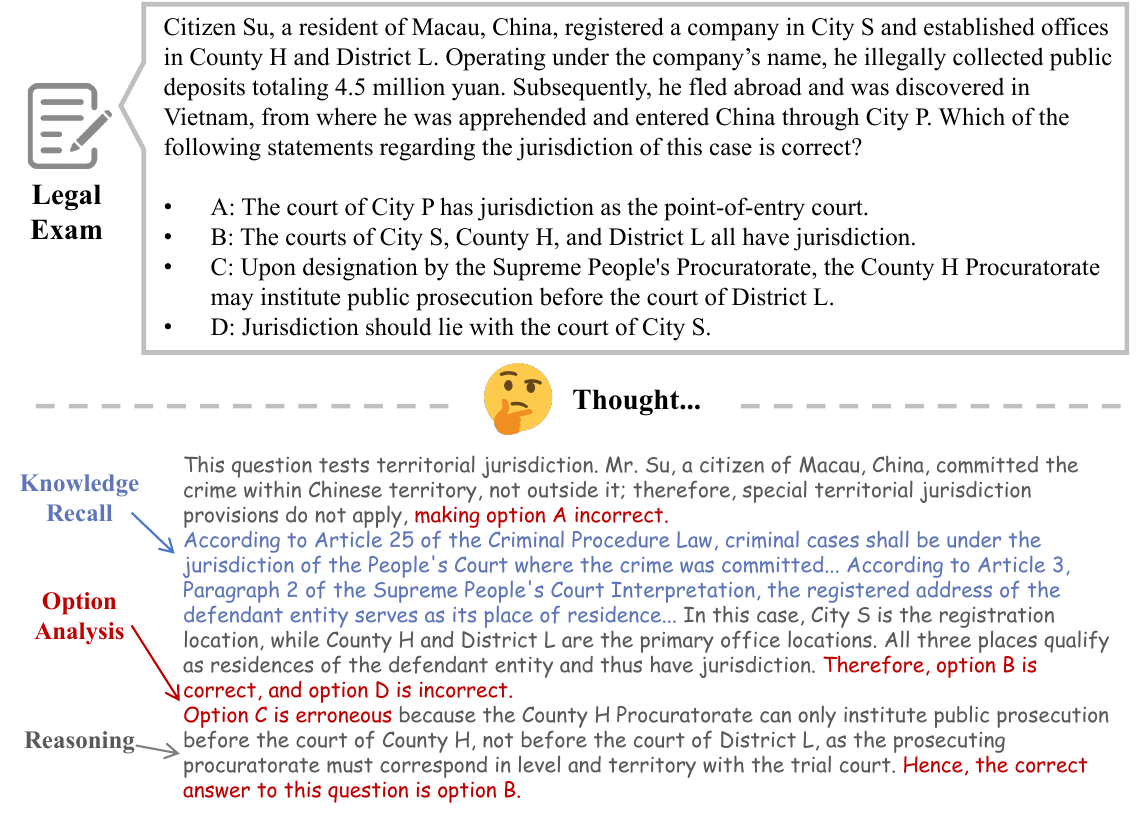}
    \caption{Illustration of a question in legal examination.}
    \label{fig:legal-exam}
\end{figure}

However, the extensive potential of the MCTS algorithm in tasks requiring complex knowledge application, sophisticated reasoning, and textual comprehension remains significantly underexplored. Figure \ref{fig:legal-exam} presents an example of legal examination, including the question statement, choices, and thoughts to the final answer \cite{zhong2019jec,yue2023disclawllm,yue2024lawllm}. The annotation of high-quality thoughts, encompassing a range of skills such as foundational legal knowledge, critical analysis of options, and comprehensive reasoning, would be prohibitively expensive. \textbf{Thus, we question whether the MCTS algorithm can yield helpful insights without much human annotation, thereby improving performance on these knowledge-intensive tasks through cost-effective supervision, in the same manner as it does for mathematics and coding.}

In this paper, we focus on the diverse challenges within the legal domain that require a broad spectrum of skills and the seamless integration of knowledge, comprehension, and reasoning. Given the diverse formats inherent in various legal tasks, they are standardized into multiple-choice questions, thereby facilitating a more convenient evaluation process.
There are several challenges of applying MCTS on the domain-specific problems. We have devoted substantial effort to enhance the integration of step-level CoTs and complicate knowledge-intensive tasks, and proposed the framework of \textbf{S}tepwise Domain \textbf{K}nowledge-Driven \textbf{R}easoning \textbf{Op}timization (SKROP).
We employ XML tags to structure thoughts and actions systematically. To aid in the comprehension and application of XML tags, we implement a preliminary warmup process before engaging in MCTS. Furthermore, we introduce a novel mechanism, called random proposal, which serves to enhance node diversity, thereby significantly expanding the search space. Given the constructed trees, SKROP generates stepwise preference pairs as training data through a meticulously devised sampling algorithm, thereby ensuring the stability of both the policy model and the value head.

However, consistently reasoning along the correct path presents a considerable challenge. Therefore, it is essential to prompt LLMs to engage in self-reflection whenever their reasoning veers off the correct route \cite{madaan2024self, zhang2024accessinggpt4levelmathematical}. \citet{qin2024o1} introduces the concept of Journey Learning, which explores supervised learning of the entire exploration path, encompassing trial-and-error and correction processes. They employ an additional LLM to generate reflective texts, serving as guiding bridges toward the target solution in response to the incorrect steps produced by the policy model.
However, potential challenges may arise in the generated refinement, particularly when the policy model fail to learn the refinement if the text diverges significantly from its expected distribution.
Therefore, we introduce the technique of \textbf{P}reference \textbf{O}ptimization towards \textbf{R}eflection \textbf{P}aths, abbreviated as PORP, with the objective of enhancing the quality of reflective texts. By increasing the probability of generating preferred reflections in contrast to those considered less appropriate, PORP aims to guide the policy model in mastering the skill of optimal reflection upon encountering missteps.

We have conducted comprehensive experiments to evaluate SKROP on complex knowledge-intensive problems, as well as the benefits of PORP. The empirical findings affirm the effectiveness of our methodologies across a variety of scenarios and analyses. Through SKROP, the policy model learns to produce deliberate reasoning steps in cooperation with domain knowledge, while also developing the capability for self-reflection with the enhancement of PORP, when it recognizes its mistakes.

Our contributions can be summarized in threefold:
\begin{itemize}
    \item Inspired by the prohibitively high cost of annotating high-quality CoTs, we propose the SKROP framework. This framework leverages the benefits of stepwise supervision generated by the MCTS algorithm, specifically adapting it for knowledge-intensive reasoning within specialized domains.
    \item We introduce PORP, which guides the policy model to generate insightful self-reflection through preference learning. This optimization process iteratively increases the probability of producing effective reflections.
    \item We analyze the contribution of the methodologies. Extensive experiments across diverse scenarios highlight the advantages of SKROP and PORP.
\end{itemize}

\section{Related Work}


\citet{wei2023chainofthoughtpromptingelicitsreasoning} introduced the concept of Chain of Thoughts, abbreviated as CoTs. CoTs are proven to be beneficial to the tasks requiring reasoning and calculation \cite{narangself,chu2024navigate}. A multitude of additional CoT-inspired methodologies have since emerged \cite{yao2024tree,besta2024got}. MCTS can generate granular, step-level supervision of thoughts, thereby enriching the training of fine-grained cognitive processes. \citet{chen2024alphamathzeroprocesssupervision} proposed AlphaMath, which bypasses the need for process annotations by leveraging MCTS, and focuses on unleashing the potential of a well-pretrained LLM to autonomously enhance its mathematical reasoning. SVPO \cite{svpo} employs MCTS to automatically annotate step-level preferences for multi-step reasoning. Furthermore, from the perspective of learning-to-rank, \citet{svpo} trained an explicit value model to replicate the behavior of the implicit reward model, complementing standard preference optimization. \citet{cpl} introduced curriculum preference learning, dynamically adjusting the training sequence of trajectory pairs in each offline training epoch to prioritize critical learning steps and mitigate over-fitting. \citet{hu-etal-2024-serts} proposed a novel retrieval method, called SeRTS, based on MCTS and a self-rewarding paradigm. Moreover, there are several other studies utilizing MCTS on reasoning, coding and planning tasks \cite{tian2024selfimprovementllmsimaginationsearching,zhang2024rest,delorenzo2024makecountllmbasedhighquality,gao2024interpretablecontrastivemontecarlo,li2024rethinkmctsrefiningerroneousthoughts,li2024elicitreasoningllmscriticguided}. Unlike prior studies, our emphasis lies in the application of the MCTS algorithm to knowledge-intensive reasoning problems, which present a multitude of challenges. Overcoming these necessitates the identification of targeted innovations to ensure effective adaptation.

However, LLMs may encounter into errors during forward reasoning. Existing research in this domain remains scarce at present. MCTSr \cite{zhang2024accessinggpt4levelmathematical} leverages systematic exploration and heuristic self-refine mechanisms to improve decision-making frameworks within LLMs. \citet{qin2024o1} introduced the concept of Journey Learning, which explores supervised learning of the entire exploration path, encompassing trial-and-error and correction processes.



\begin{figure*}
    \centering
    \includegraphics[width=\linewidth]{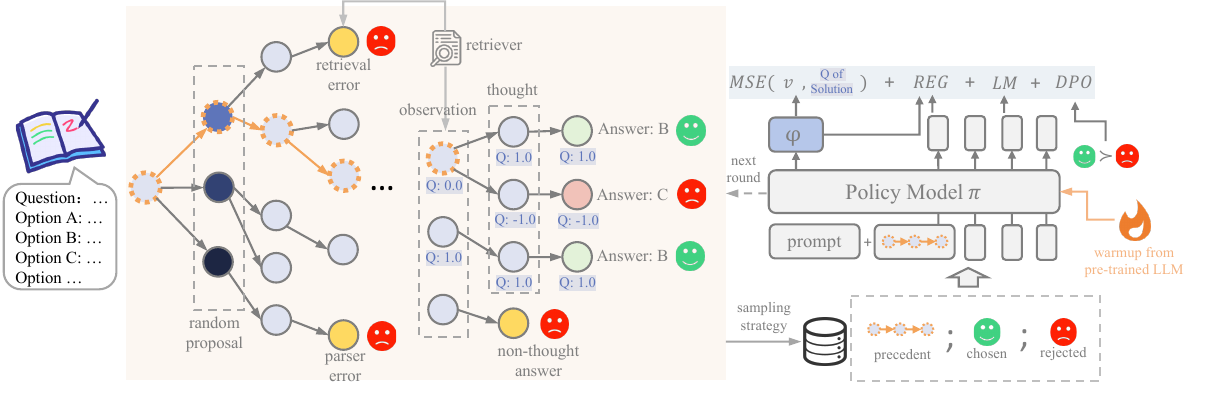}
    \caption{Framework of SKROP. SKROP builds the tree, which starts with a root node, consisting of a question and the corresponding options. The chosen (green smiling face) and rejected (red crying face) trajectories are sampled with their precedent steps, to train the policy model and the value head.}
    \label{fig:mcts}
\end{figure*}

\section{SKROP}

We formulate the knowledge-intensive reasoning problems as follows: Given a question $X$ and a list of options $C=[c_1, c_2, \dots, c_{n_c}]$, the LLM is tasked with selecting the optimal choice $\hat{A}$ as the response to the question. The response is deemed accurate solely if the predicted choice $\hat{A}$ aligns with the gold answer $A$. In our experiments, $n_c=4$.
We presents the framework of SKROP as Figure \ref{fig:mcts}. 

\subsection{MCTS}
The solution steps are organized in XML structures, including: (1) {\small \texttt{<STEP>}} is the top-level tag of the solution. (2) {\small \texttt{<PROPOSAL>}} contains a proposed answer at the beginning of search. (3) {\small \texttt{<THOUGHT>}} describes any kind of reasoning, reflection and analysis of knowledge. (4) {\small \texttt{<ACTION>}} calls the external retriever with the keyword provided by {\small \texttt{<ACTION\_INPUT>}}, then {\small \texttt{<OBSERVATION>}} wraps the retrieval results. (5) {\small \texttt{<FINAL\_ANSWER>}} contains a single option index as the answer.

We adopt a pre-trained LLM as the policy model $\pi$ to produce the XML tags. Additionally, a value head follows the last transformer layer to compute a scalar as the $Q$-value, denoted as $\varphi$. For the $t$-th step, the policy model generates the step text according to question and previous steps $y^{<t}$, similarly, the value head outputs $v^t$ of this step.
\begin{align}
y^t &= \pi ([X; C; y^{<t}])\\
v^t &= \varphi ([X; C; y^{\le t}])
\end{align}

\paragraph{Selection} 
MCTS algorithm balances the exploration and exploitation with the PUCT criterion to select the node to expand from the whole tree:
{\small \begin{equation}
    \mathrm{PUCT}(\mathbf{s}^t, \mathbf{a}^t) = Q(\mathbf{s}^t, \mathbf{a}^t) + c_{\mathrm{puct}} P_{\pi}(\mathbf{a}^t | \mathbf{s}^t) \frac{\sqrt{N_p}}{N_c + 1}
\end{equation}}
where $\mathbf{s}^t$ represents the state of $[X;C;y^{<t}]$, and $\mathbf{a}^t$ is the action to take at the $t$-th step. $Q(\mathbf{s}^t, \mathbf{a}^t)$ is the $Q$-value if taking the action $\mathbf{a}^t$ at the state $\mathbf{s}^t$, indicating the exploitation. $P_{\pi} (\mathbf{a}^t | \mathbf{s}^t)$ denotes the probability of the policy model to take the action at state $\mathbf{s}^t$. $N_p$ and $N_c$ are the number of visit times of the parent node and current node of the action respectively. $c_{\mathrm{puct}}$ is the hyper-parameter to balance these two directions. Then we can select the node following:
\begin{equation}
\begin{aligned}
    \mathbf{a}^t = \arg\max_{\mathbf{\hat{a}} \in \mathcal{A}(\mathbf{s}^t)} \mathrm{PUCT}(\mathbf{s}^t, \mathbf{\hat{a}})
\end{aligned}
\end{equation}
Here we employ $\mathcal{A}(\mathbf{s}^t)$ to represent the action space at $\mathbf{s}^t$.

\paragraph{Expansion}
The expansion process samples $n$ different steps from the distribution of the policy model by increasing the randomness.
\begin{equation}
\begin{gathered}
    \mathcal{A}(\mathbf{s}^t) = \{\mathbf{a}^t_i |\mathbf{a}^t_i \sim P_{\pi}(\mathbf{a}^t|\mathbf{s}^t),\\\mathbf{isparsable}(\mathbf{a}^t_i),i=1,2,\dots,n\}
\end{gathered}
\end{equation}
We have to ensure each action adheres to the criterion of XML, hence we use the function $\mathbf{isparsable}(\cdot)$ to drop the actions that fail to be understood. To reduce the computation cost, we merge the actions that share a BLEU-4 score larger than 0.7.

\paragraph{Simulation and Evaluation}
We employ a self-built retriever tool to search the top-$K$ related articles if it is called by the policy model\footnote{For details of the retriever, please refer to Appendix \ref{app: implementation details}.}.
For each terminal node, the reward value can be assigned as
\begin{equation}
\begin{aligned}
    r(\mathbf{s}^t, \mathbf{a}^t) = \left \{ \begin{matrix} &1.0 \quad &\text{if} \quad A=\hat{A}\\
    &-1.0 \quad &\text{otherwise} \end{matrix}\right .
\end{aligned}
\end{equation}
Assuming that $\mathbf{s}^{t+1}$ is the follow-up state of $(\mathbf{s}^t, \mathbf{a}^t)$, then
\begin{align}
    V(\mathbf{s}^{t+1}) &= (1-\lambda) \cdot v^{t+1} + \lambda \cdot r(\mathbf{s}^t, \mathbf{a}^t), \\ \lambda &= \mathbb{I}_{\mathrm{terminal}}(\mathbf{s}^{t+1})
\end{align}
where $\mathbb{I}_{\mathrm{terminal}}$ is the indicator function for terminal nodes.

\paragraph{Backpropagation}

The gold $Q$-value for training the value head is calculated as
\begin{equation}
\begin{aligned}
    Q(\mathbf{s}^t, \mathbf{a}^t) = \frac{1}{N} \sum_{\mathbf{s}^{t+i}} \mathbb{I}_{\mathbf{s}^t, \mathbf{a}^t \rightarrow \mathbf{s}^{t+i}} V(\mathbf{s}^{t+i})
\end{aligned}
\end{equation}
where $N$ denotes the number of visit times of taking this action. $\mathbf{s}^t, \mathbf{a}^t \rightarrow \mathbf{s}^{t+i}$ indicates that $\mathbf{s}^{t+i}$ is one of the subsequent states after taking action $\mathbf{a}^t$ at $\mathbf{s}^t$.

\subsection{Random Proposal}

\citet{chen2024reversethinkingmakesllms} emphasizes the concept of ``Reverse Thinking'', as it enables consistency checks between their forward and backward thinking, thereby enhancing overall reasoning performance. Inspired by their research, we introduce a new tag that has never been used in previous MCTS studies with XML structures, {\small \texttt{<PROPOSAL>}}. It indicates that the policy model proposes an answer even though the current step is far from deep consideration. The subsequent solutions will be influenced.

Additionally, we also introduce a novel mechanism called ``random proposal'', which is motivated by the requirement of diversity and exploration in MCTS algorithm. We find that if the proposals are sampled from the $\pi$, then they are most likely to point at the same option, which reduces the diversity and decrease the efficacy. Therefore, ``random proposal'' replaces the proposed option by a random sampling, i.e., $\mathbf{a}^1_i \sim \mathbf{Uniform}(C), i=1,2,\dots,n$. As shown in Figure \ref{fig:mcts}, random proposal introduce large diversity to the first level, thereby leading to different following actions during exploration.

\subsection{Warmup}

The primary challenge of start-up lies in the policy model's lack of familiarity with our customized XML tags. Exploration efforts prove futile if the step cannot be accurately parsed. Thus, warmup is essential for the general LLM to generate parsable XML steps.

Starting from a general LLM $\pi_g$, related studies often conduct pre-training and fine-tuning on a large scale of constructed XML corpus. However, it is impractical when the considerable high-quality annotated data is unavailable. Therefore, we design a two-stage warmup strategy inspired by curriculum study. Specifically, \textbf{1)} the general LLM learns only the XML tags of {\small \texttt{<STEP>}}, {\small \texttt{<PROPOSAL>}} and {\small \texttt{<FINAL\_ANSWER>}} at the first stage. At this moment, it only repeats the answers without explicit thoughts. \textbf{2)} At the second stage, we add the rest of the tags to the training data. Our experiments illustrate that this warmup approach significantly outperforms the one-way initialization gathering all tags. Detailed discussions are available in Appendix \ref{app: warmup exp}.

\begin{algorithm}[t!]
	\caption{Chosen-Rejected Pairs Sampling} 
	\label{alg:sampling} 
	\begin{algorithmic}[1]
        \REQUIRE $\Gamma, \epsilon, \delta$
        \STATE $\mathcal{D} \gets []$
        \FOR{$\tau \in \Gamma$}
            \STATE $\mathcal{P}_{\mathbf{Sb}}, \mathcal{P}_{\mathbf{SD}},\mathcal{P}_{\mathbf{O}} = [], [], []$
            \STATE $\hat{\tau} \gets \mathbf{RmNonVisited}(\tau)$ \label{line:rmnonvisited}
            \STATE $\tilde{\tau} \gets \mathbf{RmNonThought}(\hat{\tau})$ \label{line:rmnonthought}
            \STATE $\mathcal{S} = \left \{ \right \} $ \label{line:pos set}
            \FOR{$c \in \tilde{\tau}$}
                \IF{$c$ is a correct leaf node}
                    \STATE $\mathcal{S} \gets \mathcal{S} \cup \left \{c \right \} \cup \mathbf{Parents}(c)$
                \ENDIF
            \ENDFOR \label{line:endforlabel}
            \FOR{$c \in \mathcal{S}$}
                \FOR{$i = 1$ to $2$}
                \STATE Append $\tilde{c} \sim \mathbf{Sample}_{\mathrm{Sb}}(c, \delta)$ to $\mathcal{P}_{\mathbf{Sb}}$
                \ENDFOR
                \STATE Append $\tilde{c} \sim \mathbf{Sample}_{\mathrm{SD}}(c, \delta)$ to $\mathcal{P}_{\mathbf{SD}}$
                \STATE Append $\tilde{c} \sim \mathbf{Sample}_{\mathrm{O}}(c, \delta)$ to $\mathcal{P}_{\mathbf{O}}$
            \ENDFOR
            \STATE Balance $|\mathcal{P}_{\mathbf{Sb}}|: |\mathcal{P}_{\mathbf{SD}}|:|\mathcal{P}_{\mathbf{O}}|=2:1:1$ according to $\epsilon$ \label{line:balance}
            \STATE $\hat{\mathcal{D}} \gets \mathcal{P}_{\mathbf{Sb}}+ \mathcal{P}_{\mathbf{SD}}+\mathcal{P}_{\mathbf{O}}$
            \STATE Mask proposals and observations in $\hat{\mathcal{D}}$
            \STATE $\mathcal{D} \gets \mathcal{D} + \hat{\mathcal{D}}$
        \ENDFOR
        \RETURN $\mathcal{D}$
	\end{algorithmic} 
\end{algorithm}

\subsection{Chosen-Rejected Pairs Sampling}

Formally, we use $\mathcal{T}_{w}$ to represent a chosen solution, and $\mathcal{T}_l$ to represent a rejected solution, given the question and previous steps $\mathcal{T}_p$. In this way, the $Q$-value sequences of $\mathcal{T}_w$ and $\mathcal{T}_l$ can be denoted as $Q_w$ and $Q_l$ respectively. The sampling strategy of chosen-rejected pairs is shown in Algorithm \ref{alg:sampling}.

We use $\Gamma$ to represent the collection of all trees. For each tree $\tau$, the preferred nodes are only enumerated on the paths that finally lead to at least one correct {\small \texttt{<FINAL\_ANSWER>}} tag. So we collect the candidate nodes into a set $\mathcal{S}$ in Line \ref{line:pos set}-\ref{line:endforlabel}. The rejected nodes are collected considering: 1) \textbf{Siblings} of the chosen node ($\mathcal{P}_{\mathbf{Sb}}$\footnote{Short for ``\textbf{Sb}ling''.}). In this case, there is only the last-step difference within the pairs. It makes the policy model clearly learns the contribution of each single step. 2) \textbf{Non-sibling nodes of the same depth} ($\mathcal{P}_{\mathbf{SD}}$\footnote{Short for ``\textbf{S}ame \textbf{D}epth''.}). The requirement of the same depth is due to that we don't want to leave a shortcut to easily identify the chosen trajectory. Therefore we have to ensure the considerable scale of these pairs. 3) \textbf{Non-sibling nodes of different depth, i.e. the other nodes} ($\mathcal{P}_{\mathbf{O}}$), representing more general cases. Note that we apply a margin filter to all pairs, i.e. $\mathcal{T}_w \succ \mathcal{T}_l |\mathcal{T}_p, Q_w \ge Q_l + \delta$. This is because that our confidence on the $Q$-value is limited to the exploration. Close $Q$-values may be caused by unexhausted search, rather than the essential quality distinctions.
To prevent from the dataset over-concentrates on some examples with too many pairs, we employ a hyper-parameter, $\epsilon$, to denote the maximum number pairs of a single question, then balance all pairs from different source as in Line \ref{line:balance}. After collecting all pairs of three sampling sources, we merge them all as the collection for the current tree. When conducting supervised fine-tuning, the contents in proposal tags and observation tags are masked, since they are not generated by the policy model during inference. Forcing the model to fit those proposals and knowledge will introduce noise and interfere.

Additionally, considering the efficiency, our MCTS performs the simulation without rollout \cite{silver2017masteringchessshogiselfplay}. It produces some expanded nodes that have never achieved to the final answer. They are hard to be evaluated especially when the value model has not learned to fit the $Q$-values well. So we remove these non-visited nodes as in Line \ref{line:rmnonvisited}.

There is another interesting finding in our analysis. The policy model tends to generate shorter and shorter thinking steps without the control of step length. For example, the model would like to directly generate the answer without thinking and analysis, after retrieving the useful knowledge, which makes the reflection challenging and reduces the interpretability. So we add a restriction to avoid non-thought answers as in Line \ref{line:rmnonthought}. Detailed discussion about the generation of non-thought answers is provided in Appendix \ref{sec:non-thought}.

\subsection{Training}

We train the value head and policy model together with loss functions considering different perspectives. The aim of preference alignment is achieved by increasing the likelihood of generating the chosen one over the rejected one.
\begin{gather}
    \phi = \beta \log \frac{\pi_{\theta}(\mathcal{T}_w|\mathcal{T}_p)}{\pi_{\mathrm{ref}}(\mathcal{T}_w|\mathcal{T}_p)} - \beta \log \frac{\pi_{\theta}(\mathcal{T}_l|\mathcal{T}_p)}{\pi_{\mathrm{ref}}(\mathcal{T}_l|\mathcal{T}_p)} \\
    \mathcal{L}_{\mathrm{DPO}} = -\log \sigma (\phi)
\end{gather}
Considering that the precision of $Q$-values is limited by the extend of search, we train the value head with a margin, to prevent from over-fitting to the estimated value,
\begin{equation}
\begin{aligned}    
    \mathcal{L}_{\mathrm{MSE}} = & \bigg ( \max\left (0, \left (\varphi \left (\mathcal{T}_w|\mathcal{T}_p\right )-Q_w \right )^2-\gamma \right )\\+&\max\left (0, \left (\varphi \left (\mathcal{T}_l|\mathcal{T}_p \right )-Q_l \right )^2-\gamma \right ) \bigg ) \times \frac{1}{2}
\end{aligned}
\end{equation}
According to previous studies \cite{feng2024analyzingunderstandinglimitationsdpo,pal2024smaugfixingfailuremodes}, the logits of the $\mathcal{T}_w$ may descend together with $\mathcal{T}_l$ in DPO. Therefore we add the language modeling loss to avoid degradation, following
\begin{equation}
\begin{aligned}
    \mathcal{L}_{\mathrm{LM}} = -\log \pi_{\theta}(\mathcal{T}_w|\mathcal{T}_p)
\end{aligned}
\end{equation}
Inspired by \citet{svpo}, we add an additional regularization term to leverage the coherence between preference learning and value prediction and perform multi-task learning,
\begin{equation}
    \mathcal{L}_{\mathrm{REG}} = \left [\phi - \mathbf{sg}\left (v_w - v_l \right ) \right ] ^2
\end{equation}
Then the training loss can be formally written as
\begin{equation}
\begin{aligned}
    \mathcal{L} = &\mathbb{E}_{(\mathcal{T}_p, \mathcal{T}_w, \mathcal{T}_l, Q_w, Q_l)\sim \mathcal{D}} [ \mathcal{L}_{\mathrm{DPO}} \\+ &\alpha_1 \mathcal{L}_{\mathrm{MSE}} + \alpha_2 \mathcal{L}_{\mathrm{LM}} + \alpha_3 \mathcal{L}_{\mathrm{REG}} ]
\end{aligned}
\end{equation}
where $\alpha_1,\alpha_2,\alpha_3,\beta,\gamma$ are hyper-parameters to balance the loss value.

\begin{figure}[t!]
    \centering
    \includegraphics[width=\linewidth]{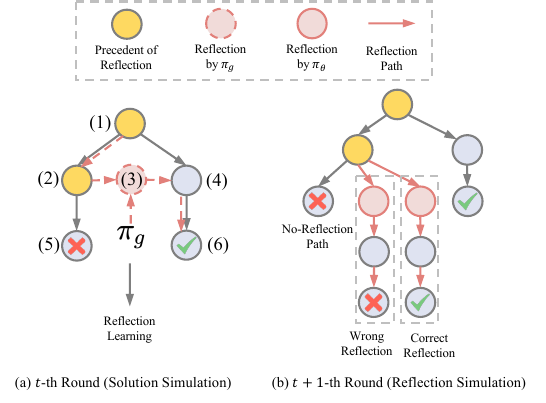}
    \caption{PORP optimizes the preference of reflection paths. The dotted lines are built for reflection training. We highlight the reflection paths in red.}
    \label{fig:porp}
\end{figure}

\section{Preference Optimization towards Reflection Paths}

Reflection makes the policy model to refine its knowledge and reasoning according to both the feedback from the environment and self-motivated thoughts. Given a pair of rejected solution and chosen solution, \citet{qin2024o1} adopts an additional LLM to guide the reflection thought generation, which serves as the corpus for further fine-tuning. However, the guidance can not ensure the quality of the reflection texts. Therefore, we introduce the Preference Optimization towards
Reflection Paths (PORP) to facilitate the learning of effective reflections.

The approach of PORP is illustrated in Figure \ref{fig:porp}. The traditional MCTS algorithm explores different one-way reasoning solutions in a forward manner. We conduct Depth-First Search (DFS) to collect pairs of $\left \langle \mathcal{T}_w, \mathcal{T}_l \right \rangle$ given the precedent steps $\mathcal{T}_p$. Formally, we use $\mathcal{\overline{T}}_w, \mathcal{\overline{T}}_l, \mathcal{\overline{T}}_p$ to denote the chosen, rejected and precedent steps. To build the dataset of reflection, we merge parts of $\mathcal{T}_l$ into $\mathcal{T}_p$, to simulate the case that the policy model has already encountered into the wrong reasoning.
\begin{gather}
\mathcal{\overline{T}}_p = \mathcal{T}_p + {\mathcal{T}_l}_{[1:i]}, 
    \mathcal{\overline{T}}_l = {\mathcal{T}_l}_{[i+1:|\mathcal{T}_l|]}\\
    \mathrm{where} \quad i \sim \mathbf{Uniform}(2, |\mathcal{T}_l|)
\end{gather}
The above segmentation strategy ensures that the trajectories are not empty. We then prompt the general LLM $\pi_g$ to generate the reflection texts following:
\begin{equation}
    \mathcal{\overline{T}}_w = \mathcal{R}(\mathcal{\overline{T}}_p, \mathcal{\overline{T}}_l, \mathcal{T}_w) + \mathcal{T}_w
\end{equation}
where $\mathcal{R}(\mathcal{\overline{T}}_p, \mathcal{\overline{T}}_l, \mathcal{T}_w)$ indicates the reflection thought generation process from $\mathcal{\overline{T}}_l$ to $\mathcal{T}_w$ given $\mathcal{\overline{T}}_p$ \footnote{To better clarify the sampling, we take the left part of Figure \ref{fig:porp} as an example. ``$(1)$'' is $\mathcal{T}_p$; ``$(2) \to (5)$'' is $\mathcal{T}_l$; ``$(4) \to (6)$'' is $\mathcal{T}_w$; ``$(1) \to (2)$'' is $\overline{\mathcal{T}_p}$; ``$(5)$'' is $\overline{\mathcal{T}_l}$; ``$(3) \to (4) \to (6)$'' is $\overline{\mathcal{T}_w}$.}. Note that the proposal step is simply skipped if $\mathcal{T}_w$ starts from it.

Additionally, we notice that the reflection would degenerate to simple reasoning, decreasing the thought length. Therefore, we add an extra consideration to the length of the chosen solution in the sampling strategy. Detailed discussion is described in Appendix \ref{app:reflection sampling}.

\begin{table*}[t!]
    \centering
    \small
    \begin{tabular}{l|ccccc|ccccc}
    \toprule
    \multirow{2}{*}{\textbf{Method}} & \multicolumn{5}{c|}{\textbf{Qwen}} & \multicolumn{5}{c}{\textbf{LLaMA}}\\
    & \textbf{JECQA} & \textbf{NJE} & \textbf{LBK} & \textbf{UNGEE} & \textbf{AVG} & \textbf{JECQA} & \textbf{NJE} & \textbf{LBK} & \textbf{UNGEE} & \textbf{AVG} \\
    \midrule
    Zero-Shot & 53.20 & 45.25 & 72.36 & 68.75 & 59.89 & 30.00 & 24.21 & 34.91 & 35.94 & 31.27\\
    ICL & 54.60	&48.60&	71.27&	64.69	&59.79&	39.00&	34.26&	50.91&	46.56	&42.68\\
    Step-by-Step & 51.20 & 45.62 & 70.55 & 65.00 & 58.09 & 39.60 & 33.15 & 45.09 & 46.25 & 41.02 \\
    Refinement & 47.80&	43.02	&71.64&	60.31	&55.69	&34.00&	29.98	&45.09&	45.00	&38.52\\
    ANS & 56.40 & 47.30 & 73.82 & 68.74 & 61.57 & 38.60 & 37.43 & 48.36 & 47.81 & 43.05 \\
    CoT + ANS & 46.20 & 40.97 & 64.00 & 66.63 & 54.45 & 36.60 & 32.03 & 45.45 & 45.94 & 40.01 \\
    ANS + CoT & 55.40 & 47.49 & 70.18 & 67.50 & 60.14 & 42.20 & 35.94 & 48.36 & \textbf{50.94} & 44.36\\
    Self-Consistency & 47.80 & 42.46 & 69.09 & 59.69 & 54.76 & 42.40 & 35.20 & 50.55 & 50.31 & 44.62\\
    RAG & 53.60 & 46.93 & 73.09 & 68.13 & 60.44 & 42.60 & 36.69 & 51.27 & 49.06 & 44.91 \\
    Distillation & 53.40 & 47.86 & 69.45 & 59.38 & 57.52 & 41.60 & 38.18 & 45.45 & 43.44 & 42.17 \\
    \midrule
    SKROP & 58.80 & 53.82 & 73.45 & \textbf{73.13} & 64.80 & 41.40 & \textbf{40.22} & 53.09 & 48.44 & \underline{45.79}\\
    \quad +SVPO & 58.40 & \underline{55.49} & 73.09 & \underline{72.50} & \underline{64.87} & \underline{44.00} & 38.55 & 51.27 & 48.44 & 45.57 \\
    \quad +CPL & 57.80 & 48.79 & 73.45 & 70.63 & 62.67 & 43.40 & 39.48 & 49.09	&\underline{49.69}	&45.42 \\
    \quad +Journey-Learning & \underline{59.00} & 54.93 & \underline{75.27} & 70.00 & 64.80 & 42.00 & 36.87 & \textbf{54.55} & 46.25 & 44.92 \\
    \rowcolor{gray!20}
    \quad +PORP & \textbf{59.20} & \textbf{55.87} & \textbf{76.73} & 71.56 & \textbf{65.84} & \textbf{44.40} & \underline{39.85} & \underline{53.82} & 49.06 & \textbf{46.78}\\
    \bottomrule
    \end{tabular}
    \caption{Results of main experiment.}
    \label{tab:main exp}
\end{table*}

\section{Experiments}

We compare SKROP with several other baselines, including inference-only prompt-based methods, supervised fine-tuning, RAG and distillation. Inspired by several related studies about MCTS in math and coding domain, we supplement several novel components to SKROP as the baselines for PORP, including SVPO \cite{svpo}, CPL \cite{cpl} and Journey-Learning \cite{qin2024o1}. Detailed baselines are described in Appendix \ref{app:baselines}.

We conduct experiments on JECQA \cite{zhong2019jec} and categories from DISC \cite{yue2023disclawllm}, which are publicly available testsets. We collect a set of training data from books and legal examinations, containing the questions, options and corresponding answers.
The best results are highlighted in \textbf{bold}, and the second best results are \underline{underlined}.
For non-XML methods, we employ regex to identify the predicted answers. If no answer can be parsed from the response, we randomly guess an option. We report the accuracy scores and the average accuracy (denoted as ``AVG'') across all datasets consistently in our experiments. For iterative methods, we repeat the training for 4 rounds and report the round with the best average accuracy.
To comprehensively assess our methods, we utilize both Qwen 7B and LLaMA 8B as base LLMs, representing scenarios with relatively extensive and limited Chinese domain knowledge, respectively. \textbf{For more details about the datasets and implementations, please refer to Appendix \ref{app: implementation details}.}

\subsection{Main Experiment}

The results of our main experiment are listed in Table \ref{tab:main exp}. There are several findings observed from the table, described as following:

\paragraph{SKROP enhances the reflection.} Based on Qwen, SKROP achieves an average accuracy of 64.80, surpassing the best baseline by 3.23 points. It outperforms other methods by 5.22 and 4.38 at least on NJE and UNGEE, respectively. When implemented on LLaMA, SKROP demonstrates superior performance on NJE and LBK. The results demonstrate that SKROP performs more effectively on Qwen compared to LLaMA, suggesting that \textbf{a more stronger base model unlocks greater potential in SKROP}. This is reasonable, as a stronger model excels in analyzing feedback and observations, thereby generating superior training data. In summary, SKROP surpasses the baselines across the majority of datasets, demonstrating the efficacy of harnessing the advantages of automatically generated CoTs.

\paragraph{PORP exhibits the advantage over other MCTS techniques.}
PORP has attained the highest accuracy scores on three datasets using Qwen, with an average result of 65.84, surpassing other components. The enhancement provided by PORP to reflection is noteworthy, particularly when compared to Journey-Learning, another reflection-oriented approach. Although PORP based on LLaMA shows less dramatic improvement, it still achieves the best overall accuracy of 46.78, outperforming other methods. Additionally, we observe only marginal improvement on UNGEE with reflection. Both Journey-Learning and Refinement even exhibit a slight decline on this dataset. This phenomenon indicates that this dataset doesn't require much thoroughgoing reasoning.

\begin{figure}[t!]
\centering
\subfigure[Qwen]{
\includegraphics[width=\linewidth]{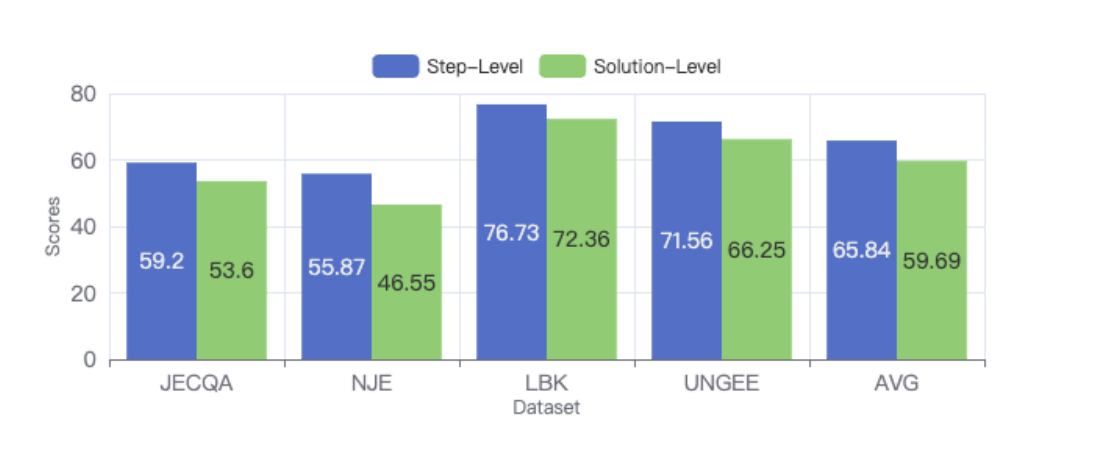} 
}
\subfigure[LLaMA]{
\includegraphics[width=\linewidth]{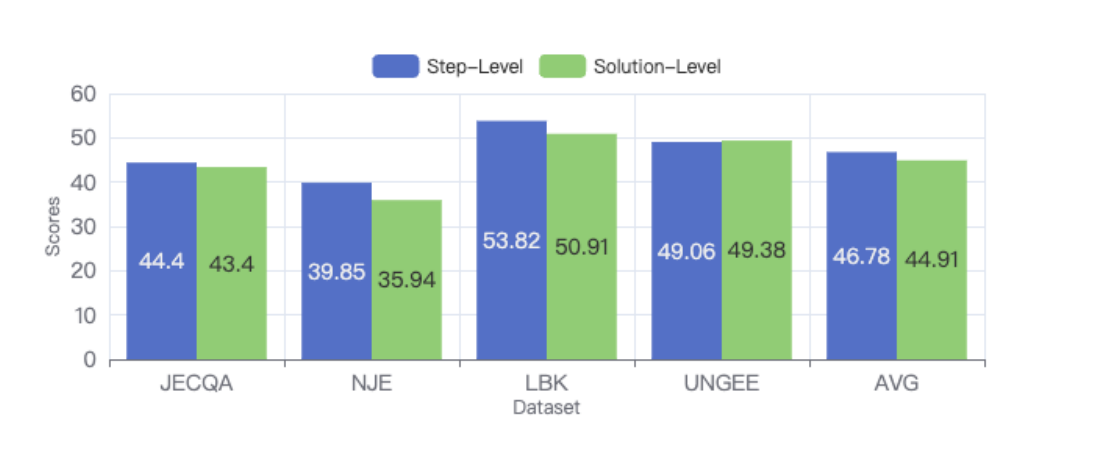} 
}
\caption{Performance of step-level and solution-level supervision.}
\label{fig:supervision granularity}
\end{figure}

\subsection{Supervision Granularity}

We compare the accuracy scores after fine-tuning on step-level and solution-level preference pairs, respectively, with the results illustrated in Figure \ref{fig:supervision granularity}. It is observed that for both LLMs, step-level supervision surpasses solution-level supervision across most datasets. Based on the Qwen model, the margin between these two granularities on the NJE dataset reaches as high as 9.32. These findings suggest that fine-grained supervision offers more detailed and significant advantages for preference learning.

\begin{table}[t!]
    \centering
    \footnotesize
    \begin{tabular}{lccccc}
    \toprule
    \textbf{Method} & \textbf{FK} & \textbf{CA} & \textbf{LR} & \textbf{LI} & \textbf{AL} \\
    \midrule
        ANS & 61.91 &	43.95&	56.08	&62.77	&50.74 \\
        CoT+ANS & 64.97&	45.22	&54.73	&65.96&	51.97\\
        RAG & 62.55&	44.59	&56.08&	57.45&	50.49\\
        SKROP & \textbf{67.26}&	50.32	&60.14&	69.14	&57.64\\
        \rowcolor{gray!20}
        \quad +PORP & 67.13&	\textbf{52.23}	&\textbf{60.81}&	\textbf{69.15}	&\textbf{58.62}\\
        \bottomrule
    \end{tabular}
    \caption{Performance of questions requiring different skills. The headers are short for ``Fundamental Knowledge'', ``Case Analysis'', ``Legal Reasoning'', ``Legal Interpretation'' and ``Application of Law'', in order. We omit ``Legal Ethics'' because there are few instances in the testset.}
    \label{tab:skills performance}
\end{table}

\subsection{Types of Skills}

To thoroughly examine the contribution of our methods, we analyze the distribution of various skill types within the test set. We categorize the questions into six types based on the required skills, as detailed in Appendix \ref{app:types of skills}.

We evaluate the accuracy of performance for each question type, comparing our method against several competitive baselines. The results are presented in Table \ref{tab:skills performance}. It is evident that SKROP significantly outperforms the baselines, particularly in tasks requiring nuanced analysis and domain-specific reasoning such as ``Case Analysis'', ``Legal Reasoning'' and ``Application of Law''. Moreover, with the incorporation of PORP, although there is a slight decrease in performance on ``Fundamental Knowledge'', the accuracy of reasoning and analysis tasks improves further. This aligns with our theoretical expectation and underscores the advantages of PORP.

\begin{figure}[t!]
    \centering
    \includegraphics[width=\linewidth]{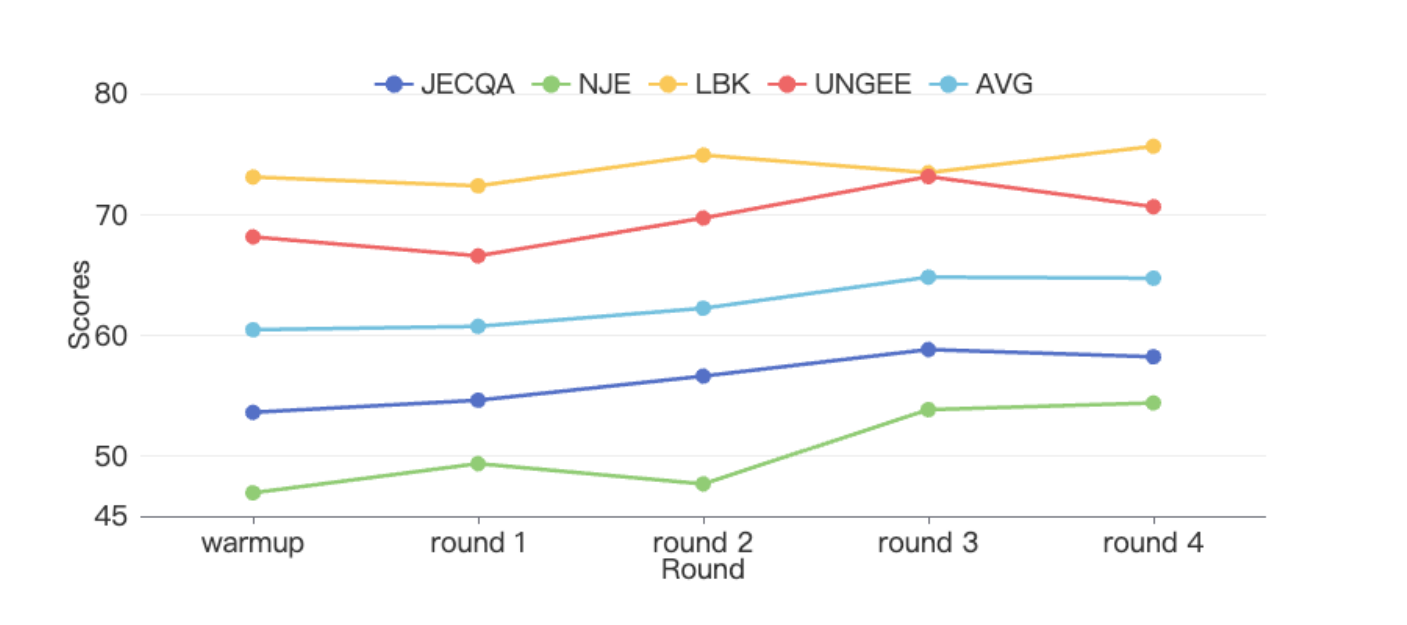}
    \caption{Accuracy scores at different rounds.}
    \label{fig:perf_over_round}
\end{figure}

\subsection{Performance over Rounds}

We illustrate the performance trend of SKROP based on the Qwen model in Figure \ref{fig:perf_over_round}. The initial state is labeled as ``warmup''. A rising trend is evident as the rounds progress. Despite occasional fluctuations, the average accuracy shows a notable enhancement compared to the warmup state. Moreover, the most substantial improvement does not occur in the first round, as the policy model seldom identifies fully correct solutions initially, resulting in less satisfactory training data and thereby impeding rapid early progress.

\begin{table}[t!]
    \centering
    \footnotesize
    \begin{tabular}{p{1.4cm}p{0.9cm}<{\centering}p{0.5cm}<{\centering}p{0.5cm}<{\centering}p{0.9cm}<{\centering}p{0.5cm}<{\centering}}
    \toprule
    \textbf{Method} & \textbf{JECQA} & \textbf{NJE} & \textbf{LBK} & \textbf{UNGEE} & \textbf{AVG}\\
    \midrule
        SKROP & 58.80 & \textbf{53.82} & \textbf{73.45} & \textbf{73.13} & \textbf{64.80} \\
        \quad w/o RP & \textbf{60.20} & 53.07 & 72.00 & 66.56 & 62.96\\
        \quad w/o LM & 55.20 & 50.28 & \textbf{73.45} & 66.56 & 61.37 \\
        \quad w/o Sb & 55.80 & 52.51 & 73.09 & 69.06 & 62.62 \\
        \quad w/o SD & 58.40 & 52.33 & \textbf{73.45} & 68.75 & 63.23 \\
    \bottomrule
    \end{tabular}
    \caption{Ablation study. ``PR'' denotes the random proposal mechanism. ``LM'' is the language modeling loss. ``Sb'' and ``SD'' denote the sibling nodes and non-sibling nodes at the same depth, respectively.}
    \label{tab:ablation}
\end{table}

\subsection{Ablation Study}

We eliminate the random proposal mechanism, language modeling loss, and two sampling sources, with the ablation results presented in Table \ref{tab:ablation}. Removing the random proposal mechanism reduces the average accuracy from 64.80 to 62.96. Without the language modeling loss, it drops by 3.43. Sampling from sibling nodes appears to be more crucial than non-sibling nodes, as they offer a more precise reward gap corresponding to a single step.

\section{Conclusion}

In this paper, we investigate the stepwise preference learning for domain knowledge-driven reasoning optimization utilizing MCTS algorithm, and propose the framework of SKROP. Additionally, we have introduced PORP and designed specific sampling strategy to improve the reflection. We have conducted extensive experiments to evaluate the advantages of our methodologies. Empirical results demonstrate the effectiveness on various legal-domain problems. 

\section*{Limitations}

While our approach is language-agnostic, our experiments primarily concentrate on the Chinese language. Moreover, our methodology possesses the theoretical flexibility to be adapted to any other domain. However, we omit empirical studies for other domains due to the absence of a comprehensive knowledge base specific to them. This presents an intriguing avenue for future research, ideally pursued by scholars equipped with extensive professional expertise and corpora in their respective domains.

\section*{Ethical Consideration}

In this paper, we aim to enhance the performance of LLMs within the legal domain through the use of automatically generated Chain of Thoughts. For experiments, we use public base models and employ public datasets as test problems. We obey the license of related works to conduct analysis. Leveraging a specialized legal LLM assistant, legal professionals and experts can significantly improve their work efficiency. Additionally, these models facilitate legal education for the general public.

\bibliography{custom}

\appendix

\section{Implementation Details}
\label{app: implementation details}

\paragraph{MCTS}
We set $c_{\mathrm{puct}}=1.5$. We sample $n=4$ responses with temperature $1.0$, using the vLLM framework \cite{kwon2023efficient}. We set the maximum depth as 16 within 40 times simulations, while increasing to 80 times simulations at most for reflection generation, since reflection requires much longer thought steps. The search is done when there are no more unexplored nodes. We have constructed a large-scale legal domain knowledge base of various articles to provide knowledge augmentation for the LLMs. Our database encompasses 2.7 million legal provisions, covering virtually all public legal documents in China. The retrieval engine employs a multi-channel strategy, incorporating both keyword and vector-based retrieval methods. The vectorization model is trained using a two-stage process based on the CoROM dual-tower framework. Additionally, a general text embedding (GTE) model has also been trained to rerank the top 1000 related results. $K$ is set to 3 in our experiment.

\paragraph{Sampling}
When sampling preference pairs for CoTs generation, we set the maximum number of pairs for each question $\epsilon = 20, \delta=0.1$. In this way, we will collect approximately 35k pairs at the last several rounds of simulations. For PORP, we balance the normal reasoning pairs and reflection pairs at the same scale to prevent the policy model from over-fitting to infinite self-reflections. During reflection sampling, we assign a weight of 0.2 to the length of reasoning steps\footnote{See Appendix \ref{app:reflection sampling}.} and a weight of 1.0 to the value gap between positive and negative pairs. The instances are then sorted and truncated according to $\epsilon$.

\paragraph{Training}
We set $\beta=0.1, \gamma=0.1$. To balance the loss items, $\alpha_1=0.25, \alpha_2=5.0, \alpha_3=0.001$. When conducting the DPO training, we set the policy model of the previous round as $\pi_{\mathrm{ref}}$.
For Qwen LLM, we use Qwen1.5-7B-Chat model\footnote{https://huggingface.co/Qwen/Qwen1.5-7B-Chat}. We employ lower-version LLMs to ensure that questions from the public testset are excluded from the model's training data, thereby circumventing spuriously high accuracy resulting from data leakage. For LLaMA model, we use the fine-tuned Chinese LLaMA by \citet{shenzhi_wang_2024}, and download the parameters\footnote{https://huggingface.co/shenzhi-wang/Llama3-8B-Chinese-Chat}.
To reduce the memory utilization, we adopt bf16, and train the models using LoRA \cite{hu2021lora} on all linear layers, with rank 16 and batch size 32. The learning rate is $10^{-5}$, optimized with a cosine scheduler. We train the models with 4 A100 80G GPUs. 

\paragraph{Datasets}
\label{app:dataset}
We list the dataset categories in Table \ref{tab:dataset}. Note that our testset covers various formats of legal questions, such as case analysis and knowledge QA. Additionally, it also includes criminal, civil, and administrative causes. Although the datasets we employ are in Chinese, the task we investigate is language-agnostic, rendering our experimental results generalizable to other languages.

\begin{table}[t!]
    \centering
    \footnotesize
    \begin{tabular}{ccm{4.5cm}}
    \toprule
    \textbf{Set} & \textbf{Scale} & \textbf{Description} \\
    \midrule
    Train & 2000 & Self-constructed training dataset\\
    JECQA & 500 & Legal-domain knowledge driven QA \\
    NJE & 537 &  National Judicial Examination \\
    LBK & 275 & Legal Basic Knowledge \\
    UNGEE & 320 & Unified National Graduate Entrance Examination\\
    \bottomrule
    \end{tabular}
    \caption{Information of the used datasets in our experiments.}
    \label{tab:dataset}
\end{table}

\paragraph{Baselines}
\label{app:baselines}
\textbf{1) Zero-Shot} \cite{xian2017zero}, which prompts the general LLM to answer the questions directly. \textbf{2) In-Context Learning (ICL)} \cite{radford2019language}, asking the model to answer the questions given one demonstration within the context. \textbf{3) Step-by-Step} \cite{kojima2022large}. A special prompt, ``Let's think step by step'', brings significant performance improvement to the LLMs. \textbf{4) Refinement} \cite{madaan2024self} asks the model to revise its response given the feedback towards the previous answer. \textbf{5) Supervised fine-tuning} on the CoTs. We prompt the general LLM to explain the question and write the reasoning thoughts. Then we pose the CoTs at different positions around the final answer to perform the fine-tuning. ``CoT + ANS'' indicates that the answer is assigned as the end of the thoughts, while ``ANS + CoT'' represents the opposite. ``ANS'' denotes the fine-tuning without thoughts. \textbf{6) Self-consistency} \cite{narangself} selects the optimal answer by aggregating votes from multiple candidate thoughts, each sampled from the same fine-tuned reasoning model. \textbf{7) RAG} \cite{lewis2020retrieval} leverages external knowledge bases to enhance response generation in agent behavior. The retriever identifies the top-$K$ relevant articles whenever the model activates the tool. By doing so, the system seamlessly incorporates both query-rewriting \cite{ma-etal-2023-query} and answer-rewriting functionalities. \textbf{8) Distillation}. We attempt to distill the knowledge and reasoning ability from larger LLMs to the smaller LLMs using black-box distillation. We adopt Qwen-Max\footnote{https://qwenlm.github.io/blog/qwen-max-0428}, which is 200B LLM, as the teacher model to generate the reasoning chain, which serves as the training data to fine-tune the student model.
We compare the following typical techniques with PORP: \textbf{9) SVPO} \cite{svpo} adds an additional loss item to maximize the value gap between chosen and rejected pairs within a pre-defined margin. \textbf{10) CPL} \cite{cpl} (Curriculum Preference Learning), dynamically adjusts the training sequence of trajectory pairs in each offline training epoch to prioritize critical learning steps and mitigate over-fitting. \textbf{11) Journey-Learning} \cite{qin2024o1} explores supervised learning of the entire exploration path, encompassing trial-and-error and correction processes.

\begin{table}[t!]
    \centering
    \scriptsize
    \begin{tabular}{p{0.7cm}<{\centering}cccccc}
    \toprule
    \textbf{Warmup} & \textbf{Round} & \textbf{JECQA} & \textbf{NJE} & \textbf{LBK} & \textbf{UNGEE} & \textbf{AVG} \\
    \midrule
    \multirow{4}{*}{\Checkmark} & Warmup & 42.60&	36.69&	51.27&	49.06&	44.91\\
         & 1 & \textbf{42.80}	&38.73&	50.91&	\textbf{49.69}	&45.53\\
         & 2 & 41.40&	\textbf{40.22}&	\textbf{53.09}&	48.44&	\textbf{45.79}\\
         & 3  & 41.40	&37.62	&52.36&	45.63&	44.25\\
    \midrule
    \multirow{4}{*}{\XSolidBrush} & 1 & 39.60&	38.92&	48.00	&41.56&	42.02\\
    & 2 & 38.40&	40.60&	46.91	&44.06&42.49\\
    & 3 & 40.40&	42.83&	46.91&	46.25&	44.10\\
    & 4 & 39.80	&40.41&	46.91	&44.69&42.95\\
    \bottomrule
    \end{tabular}
    \caption{Ablation of warmup at the start-up phase.}
    \label{tab:warmup}
\end{table}

\begin{table*}[t!]
    \centering
    \footnotesize
    \begin{tabular}{m{5cm}lm{5cm}}
    \toprule
    \textbf{Question} & \textbf{Type} & \textbf{Reason}\\
    \midrule
        \textit{According to Article XX of the Civil Code, which of the following is a necessary condition for the formation of a contract?} & Fundamental Knowledge & The question tests memorization and understanding of a provision in the Civil Code. \\
        \midrule
        \textit{Party A and Party B entered into a sales contract. Party A failed to make the payment as agreed. How should Party B assert their rights?} & Case Analysis & The question provides a specific case and requires analyzing legal relationships and solving the issue. \\
        \midrule
        \textit{According to Article XX of the Criminal Law, does Party A's behavior constitute a crime?} & Legal Reasoning & The question requires using the Criminal Law provision to perform logical reasoning. \\
        \midrule
        \textit{Explain the meaning of the ``principle of good faith'' in Article XX of the Civil Code.} & Legal Interpretation & The question requires interpreting the meaning of a legal provision. \\
        \midrule
        \textit{During the Kaiyuan era of Emperor Xuanzong's reign, a villager named Zhang from Xu Prefecture in Henan Circuit went hunting and spotted a pheasant in the woods. He drew his bow and shot an arrow, but unfortunately hit Li, a herbalist, who was gathering herbs, killing him with an arrow to the head. According to Tang Dynasty law, what crime would Zhang's action constitute?} & Application of Law & The question requires applying a legal provision to a specific scenario. \\
        \midrule
        \textit{Does the lawyer's behavior comply with professional ethics?} & Legal Ethics & The question tests understanding of legal professional ethics.\\
        \bottomrule
    \end{tabular}
    \caption{Types of skills in the testset.}
    \label{tab:skills info}
\end{table*}

\section{Warmup}
\label{app: warmup exp}

We conducted an ablation study by eliminating the warmup phase at the initial stage, with the results detailed in Table \ref{tab:warmup}. To ensure fairness, we executed four iterations for each setting. It is observed that the warmup phase notably outperformed all four rounds without warmup. The subsequent rounds following warmup present the upward trend, achieving the highest overall accuracy of 45.79 in the second round. This finding highlights the essential role of the warmup phase.


\begin{figure}[t!]
    \centering
    \includegraphics[width=\linewidth]{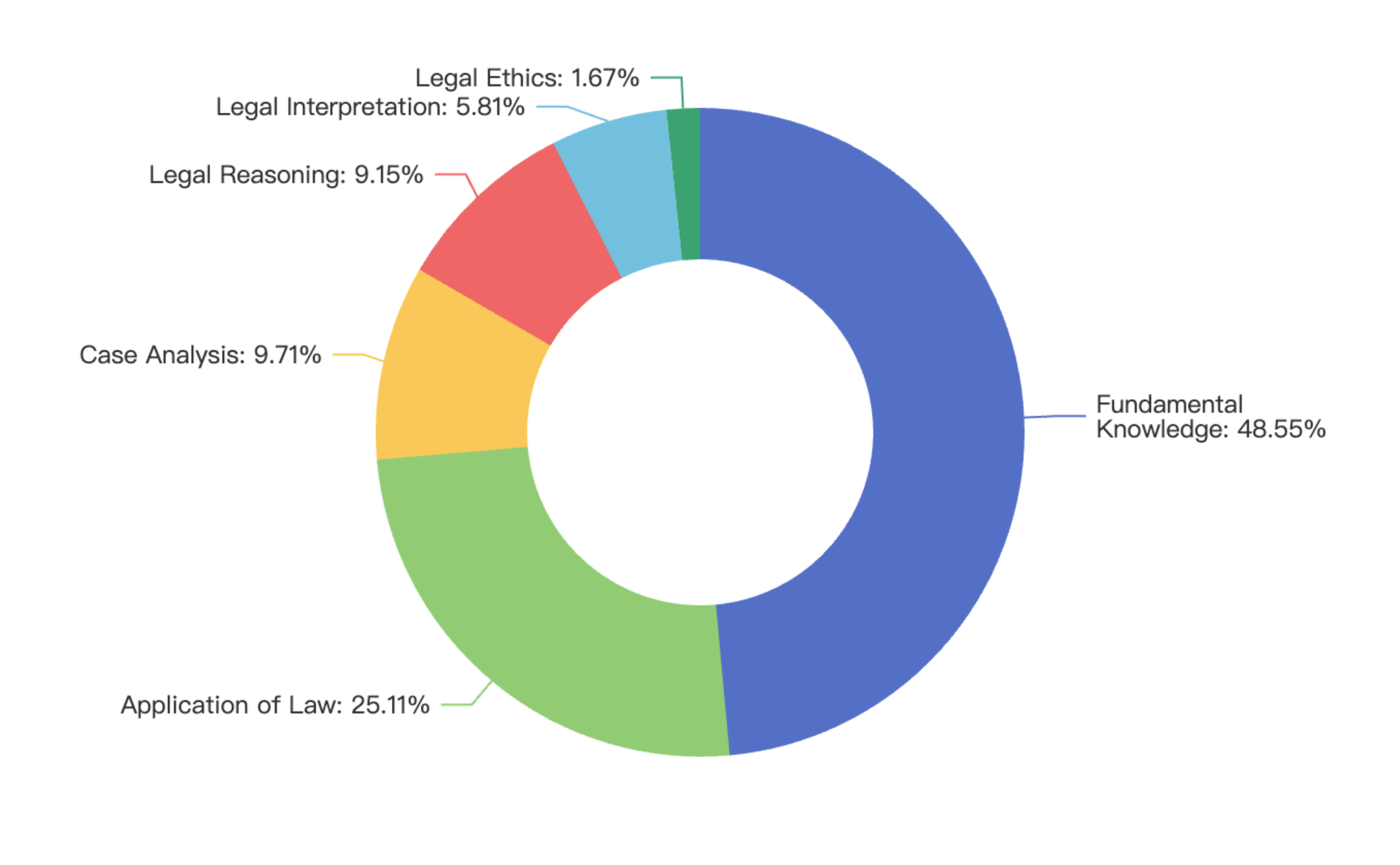}
    \caption{Distribution of skill types.}
    \label{fig:skill type distribution}
\end{figure}

\section{Types of Skills}
\label{app:types of skills}

To thoroughly examine the contribution of our methods, we analyze the distribution of various skill types within the test set. We categorize the questions into six skill types. We employ larger model to label each problem with the type of required skill with the criterion shown in Table \ref{tab:skills info}. The corresponding distribution is illustrated in Figure \ref{fig:skill type distribution}. Our observations reveal that nearly half of the questions pertain to fundamental legal knowledge. Due to the amount of ``Legal Ethics'' is too small, we omit this category when reporting the performance.

\begin{figure}[t!]
    \centering
    \includegraphics[width=\linewidth]{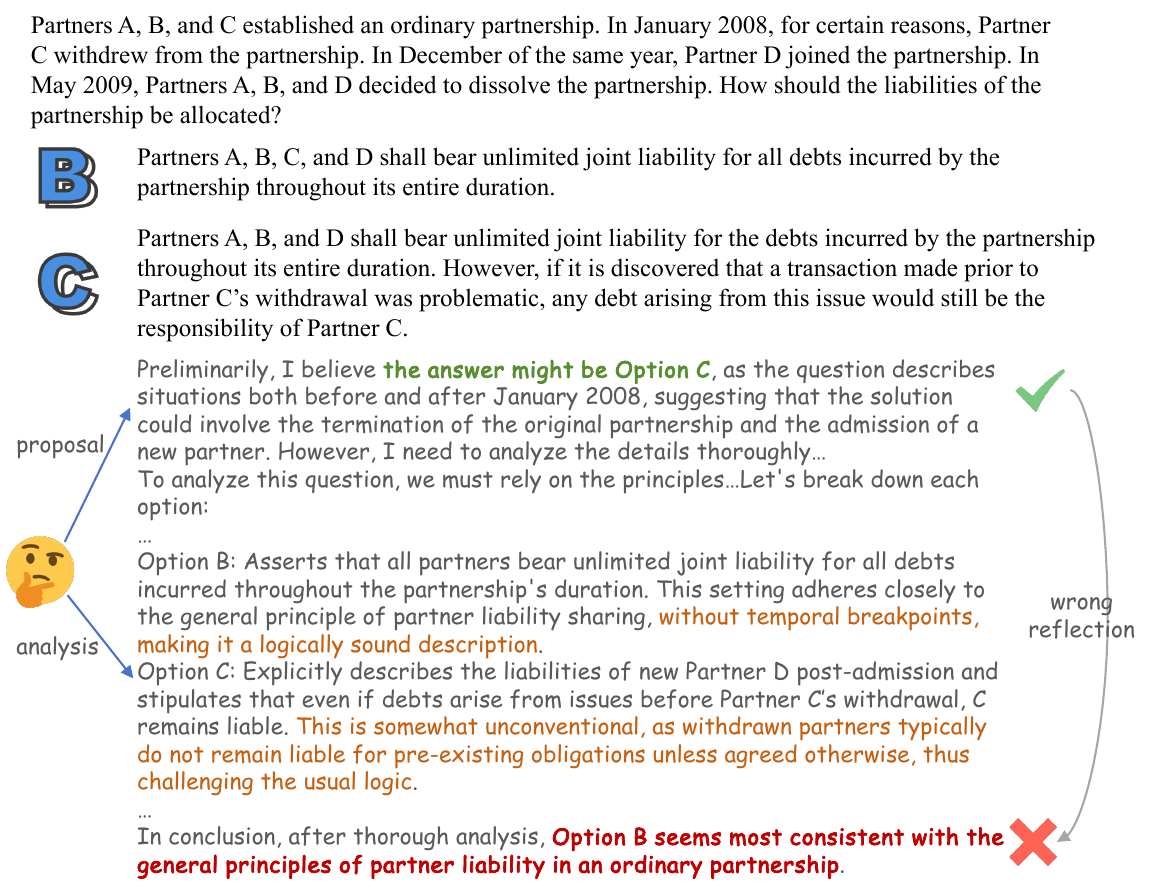}
    \caption{An instance where the LLMs transform the accurate proposal into erroneous responses, indicating a potential lack of confidence in their knowledge and reasoning abilities.}
    \label{fig:correct-porposal-to-wrong-answer}
\end{figure}

\section{Findings from Main Experiment}

\paragraph{CoTs are not always beneficial.}
We position the CoTs at various locations to conduct supervised fine-tuning. Surprisingly, directly aligning with the gold options yields the highest overall accuracy score of 61.57 on Qwen. On LLaMA, however, the ``ANS + CoT'' approach slightly outperforms ``ANS''. Their success mirrors the efficacy of SKROP's proposal mechanism, as both strategies provide the answer before explaining their thoughts. The ``CoT + ANS'' method demonstrates suboptimal performance for both base LLMs, likely due to potential noise within the CoTs.

\paragraph{Model gap hinders the performance of distillation.}
When we endeavor to distill reasoning capabilities from larger LLM (specifically, Qwen-Max 200B) using automatically annotated thoughts, the resulting accuracy proves less remarkable than that achieved through self-distillation. The average accuracy falls 2.62 points below ``ANS + CoT'' on Qwen and 2.19 points below on LLaMA.

\paragraph{Prompt-based tricks are unstable for domain knowledge-driven reasoning.}
We have explored prompt-based techniques for our task, which have demonstrated utility in mathematical calculations and coding. However, their effectiveness appears inconsistent. Compared to ``Zero-Shot'' on Qwen, ``ICL'' enhances accuracy from 45.25 to 48.60 on NJE but reduces it by 4.06 on UNGEE. Both ``ICL'' and ``Step-by-Step'' significantly boost the average accuracy for LLaMA relative to ``Zero-Shot'', whereas they exert negative impact on Qwen.

\section{Self-Refinement}
\label{sec:self-refine}

We analyze the flaws of self-refinement, after observing its unsatisfactory accuracy during experiments. We find that the thoughts and reasoning are untrustworthy without specific supervision. We illustrate an example of wrong reflection in Figure \ref{fig:correct-porposal-to-wrong-answer}. In the example, the model initially suggests the correct answer. However, during the reasoning process, it generates vague and ambiguous thoughts and judgments, which confuse the inference and ultimately result in an incorrect final answer. This failure of prompt-based refinement underscores the necessity for high-quality training data and rigorous supervision to produce robust and trustworthy knowledge-driven reasoning.

\begin{figure}[t!]
    \centering
    \includegraphics[width=0.6\linewidth]{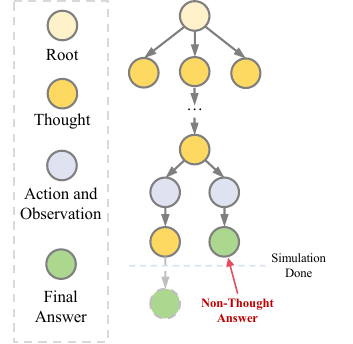}
    \caption{Generation of non-thought answers during iterations.}
    \label{fig:non-thought answer}
\end{figure}

\begin{figure}[t!]
    \centering
    \subfigure[w/o Length Restriction]{
    \includegraphics[width=\linewidth]{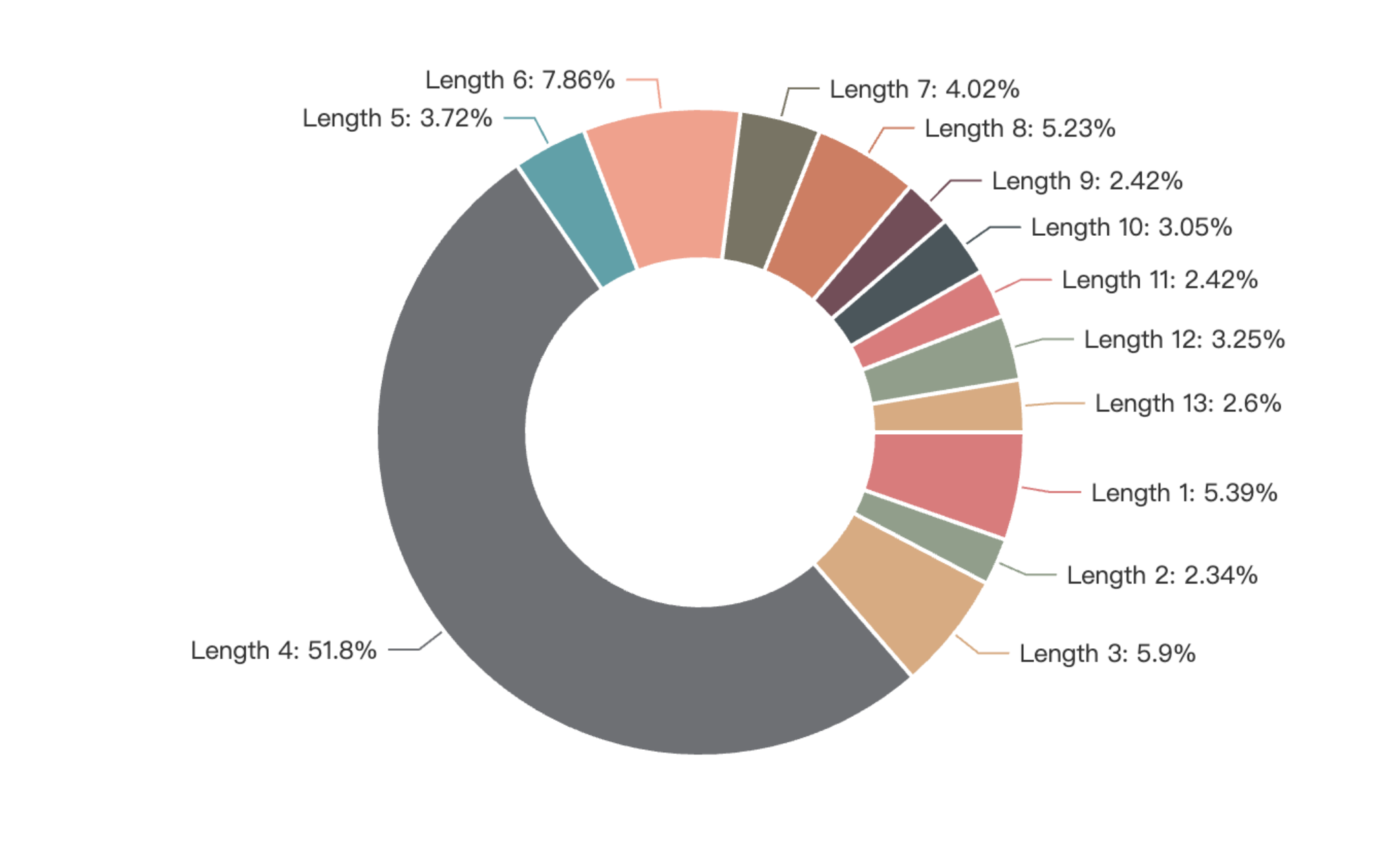}
    \label{fig:w/o length strict}
    }
    \subfigure[w Length Restriction]{
    \includegraphics[width=\linewidth]{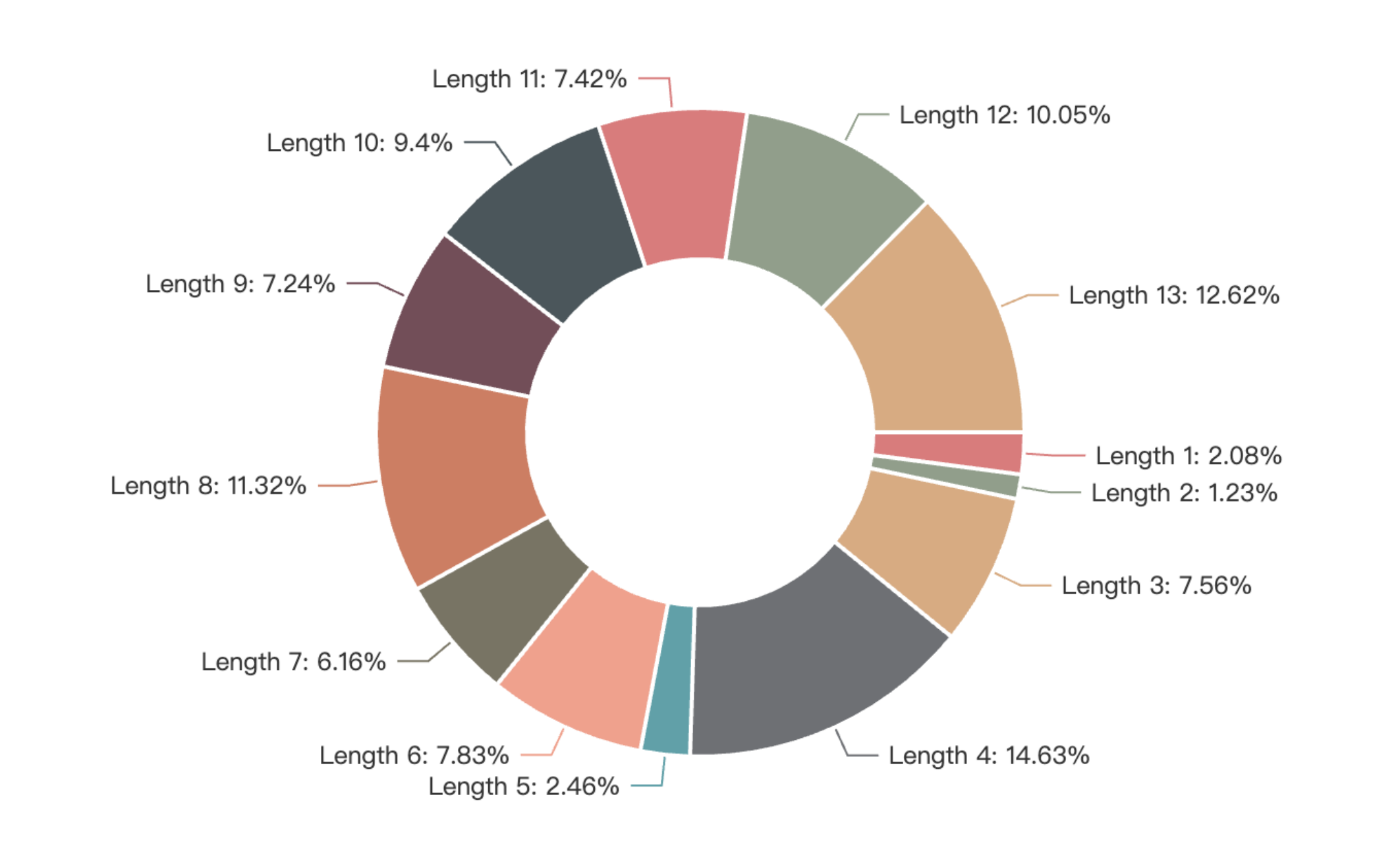}
    \label{fig:w length strict}
    }
    \caption{Length restriction in PORP sampling.}
    \label{fig:porp sampling}
\end{figure}

\section{Non-Thought Answer}
\label{sec:non-thought}

We have observed an intriguing pattern during the experiments conducted with SKROP. Specifically, we noted that \textbf{the policy model tends to directly output the final answer tag after invoking the retriever}. We analyzed the underlying reasons for this phenomenon and illustrated our findings in Figure \ref{fig:non-thought answer}. Compared to the standard path ({\small \texttt{<OBSERVATION>}} $\to$ {\small \texttt{<THOUGHT>}} $\to$ {\small \texttt{<FINAL\_ANSWER>}}), the non-thought answer ({\small \texttt{<OBSERVATION>}} $\to$ {\small \texttt{<FINAL\_ANSWER>}}) is more concise, thereby being explored earlier within the limited simulation times. Consequently, this results in a higher prevalence of such nodes in the training data. However, this phenomenon is not expected, since intermediate answer lacks medium 
We have observed an intriguing pattern during the experiments conducted with SKROP. Specifically, we noted that \textbf{the policy model tends to directly output the final answer tag after invoking the retriever}. We analyzed the underlying reasons for this phenomenon and illustrated our findings in Figure \ref{fig:non-thought answer}. Compared to the standard path ({\small \texttt{<OBSERVATION>}} $\to$ {\small \texttt{<THOUGHT>}} $\to$ {\small \texttt{<FINAL\_ANSWER>}}), the non-thought answer ({\small \texttt{<OBSERVATION>}} $\to$ {\small \texttt{<FINAL\_ANSWER>}}) is more concise, thereby being explored earlier within the limited simulation times. Consequently, this results in a higher prevalence of such nodes in the training data. However, this phenomenon is undesirable because immediate answers lack the connecting thoughts that explain how observations lead to the consequent answer, thereby hindering the interpretability of the solution. Therefore, we add an additional restrict during training data sampling, to remove the non-thought answers from the set of candidate chosen solutions, as shown in Line \ref{line:rmnonthought} of Algorithm \ref{alg:sampling}.

\section{Reflection Sampling}
\label{app:reflection sampling}

To ensure the quality of training data in self-reflection, we extended the simulation times and maximum reasoning steps. We observed that most solutions still emerge within a limited number of steps, rarely utilizing reflection, as depicted in Figure \ref{fig:w/o length strict}. Consequently, we revised the sampling strategy to prioritize longer reasoning texts. Specifically, we incorporated length as a weighted factor during sampling. By introducing this adjustment, the distribution of reasoning steps became more balanced, as illustrated in Figure \ref{fig:w length strict}.

\section{XML Example}

We present an example of XML tag in Table \ref{tab:xml example}.

\begin{table*}[p]
    \centering
    \small
    \begin{tabular}{m{15cm}}
    \toprule
         <Question> Country A and Country B have disputes over the Exclusive Economic Zone (EEZ): Country A believes it has territorial sovereignty over the EEZ, and thus aircraft from Country B must obtain permission from Country A to fly over or navigate through. According to the United Nations Convention on the Law of the Sea, which of the following statements is correct?\\
A: Country A's claim is incorrect; it does not possess territorial sovereignty over the EEZ but enjoys certain sovereign rights as stipulated by the convention.\\
B: Country B has the freedom of navigation, overflight, laying of submarine cables, and exploitation of biological resources in Country A's EEZ.\\
C: Country A can exercise a certain degree of control over specific issues such as customs, fiscal affairs, immigration, or health within its EEZ.\\
D: The rights of Country A over the EEZ are inherent, including the airspace above and the seabed below.</Question>\\
\midrule
         <step> <proposal>I guess the answer might be A.</proposal> </step>\\<step> <thought>Understanding the question content: We need to focus on the definition of the EEZ under the United Nations Convention on the Law of the Sea and the various rights involved.</thought> <action>retriever</action> <action\_input>Exclusive Economic Zone Territorial Sovereignty</action\_input> <observation>The Exclusive Economic Zone (EEZ) is an area beyond and adjacent to the territorial sea of a coastal state, subject to its sovereignty for the purpose of exploring and exploiting, conserving, and managing natural resources, both living and non-living, of the seabed and subsoil and superjacent waters, and with regard to other activities for the economic exploitation and exploration of the zone, such as the production of energy from the water, currents, and winds. However, the coastal state does not have territorial sovereignty over the EEZ.</observation> </step> \\<step> <thought>Analyzing option A: Option A states that Country A does not have territorial sovereignty over the EEZ but enjoys certain rights, which aligns with the convention, so A is correct.</thought> </step> \\<step> <thought>Continuing to analyze option B: Activities of Country B in Country A's EEZ need to comply with the laws of the coastal state and international agreements. While enjoying certain freedoms, the statement in option B is not entirely accurate.</thought> </step>\\ <step> <thought>Analyzing option C: Although Country A can indeed exercise control over certain specific matters, this does not equate to territorial sovereignty, making option C incorrect.</thought> </step>\\ <step> <thought>Analyzing option D: Rights over the EEZ are based on international conventions rather than being inherent, hence option D is inaccurate.</thought> </step>\\ <step> <final\_answer>A</final\_answer> </step>\\
         \bottomrule
    \end{tabular}
    \caption{Example of XML tags.}
    \label{tab:xml example}
\end{table*}


\end{document}